\documentclass[journal]{IEEEtran}

\usepackage{amsmath,amsfonts}
\usepackage{array}
\usepackage[caption=false,font=normalsize,labelfont=sf,textfont=sf,position=top]{subfig}
\usepackage{textcomp}
\usepackage{stfloats}
\usepackage{url}
\usepackage{verbatim}
\usepackage{booktabs}
\usepackage{orcidlink}
\usepackage{graphicx}
\def\BibTeX{{\rm B\kern-.05em{\sc i\kern-.025em b}\kern-.08em
    T\kern-.1667em\lower.7ex\hbox{E}\kern-.125emX}}
\usepackage{balance}
\usepackage{enumitem}

\hypersetup{
    colorlinks=true,
    linkcolor=purple,
    filecolor=cyan,      
    urlcolor=teal,
    citecolor=blue,
    }

\usepackage{algorithm}
\usepackage[noend]{algpseudocode}
\let\Algorithm\algorithm
\renewcommand\algorithm[1][]{\Algorithm[#1]}

\algrenewcommand{\algorithmiccomment}[1]{\hskip3px$\#$ #1}
\algdef{SE}[SUBALG]{Indent}{EndIndent}{}{\algorithmicend\ }
\algtext*{Indent}
\algtext*{EndIndent}

\usepackage{tikz}
\usepackage{tikzsymbols}
\usetikzlibrary{shapes,positioning,calc,arrows,arrows.meta,circuits.logic.US}

\usepackage[square,numbers,sort&compress]{natbib}
\renewcommand{\cite}{\citep}

\usepackage{listings}
\usepackage{xcolor}
\definecolor{codegreen}{rgb}{0,0.6,0}
\definecolor{codegray}{rgb}{0.3,0.3,0.3}
\definecolor{codepurple}{rgb}{0.5,0,0.8}
\definecolor{backcolour}{rgb}{0.97,0.97,0.97}
\lstdefinestyle{mystyle}{
    backgroundcolor=\color{backcolour},   
    commentstyle=\color{codegreen},
    keywordstyle=\color{magenta},
    numberstyle=\tiny\color{codegray},
    stringstyle=\color{codepurple},
    basicstyle=\ttfamily\scriptsize,
    breakatwhitespace=false,         
    breaklines=true,                 
    captionpos=b,                    
    keepspaces=true,                 
    numbers=none,                    
    numbersep=5pt,                  
    showspaces=false,                
    showstringspaces=false,
    showtabs=false,                  
    tabsize=2
}
\begin{document}
\title{Fitness Approximation through Machine Learning\\
{\normalsize\textbf{\color{purple} Submitted to Special Issue on Machine Learning Assisted Evolutionary Computation}}}


\author{Itai~Tzruia~\orcidlink{0009-0004-5019-017X},
    Tomer~Halperin~\orcidlink{0009-0009-9050-8118},
    Moshe~Sipper~\orcidlink{0000-0003-1811-472X},
    Achiya~Elyasaf~\orcidlink{0000-0002-4009-5353}
\IEEEcompsocitemizethanks{%
\IEEEcompsocthanksitem A. Elyasaf is with the Software and Information Systems Engineering Department, Ben-Gurion University of the Negev, Israel.\protect\\
E-mail: \texttt{achiya@bgu.ac.il}
\IEEEcompsocthanksitem I. Tzruia, T. Halperin, and M. Sipper are with the Computer Science Department, Ben-Gurion University of the Negev, Israel.\protect\\
E-mail: \texttt{itaitz@post.bgu.ac.il}, \texttt{tomerhal@post.bgu.ac.il}, and \texttt{sipper@bgu.ac.il}}
}

\markboth{IEEE Transactions on Evolutionary Computation}{Tzruia et al.}

\maketitle

\begin{abstract}
We present a novel approach to performing fitness approximation in genetic algorithms (GAs) using machine-learning (ML) models, through dynamic adaptation to the evolutionary state.
Maintaining a dataset of sampled individuals along with their actual fitness scores, we continually update a fitness-approximation ML model throughout an evolutionary run.
We compare different methods for:
1) switching between actual and approximate fitness, 
2) sampling the population, and 
3) weighting the samples. 
Experimental findings demonstrate significant improvement in evolutionary runtimes, with fitness scores that are either identical or slightly lower than that of the fully run GA---depending on the ratio of approximate-to-actual-fitness computation. 
Although we focus on evolutionary agents in Gymnasium (game) simulators---where fitness computation is costly---our approach is generic and can be easily applied to many different domains.
\end{abstract}


\section{Introduction}
A genetic algorithm (GA) is a population-based metaheuristic optimization algorithm that operates on a population of candidate solutions, referred to as individuals, iteratively improving the quality of solutions over generations. GAs employ selection, crossover, and mutation operators to generate new individuals based on their fitness values, computed using a fitness function \citep{558db5e0-7a5a-3da1-8b88-2f13b93dbaf1}.

GAs have been widely used for solving optimization problems in various domains, such as telecommunication systems \citep{jha2018energy}, energy systems \citep{mayer2020environmental}, and medicine \citep{hemanth2019modified}.
Further, GAs can be used to evolve agents in game simulators. For example, \citet{GARCIASANCHEZ2020105032} employed a GA to enhance agent strategies in Hearthstone, a popular collectible card game, and 
\citet{Elyasaf2012EvolutionaryDesign} evolved top-notch solvers for the game of FreeCell.

An accurate evaluation of a fitness function is often computationally expensive, particularly in complex and high-dimensional domains, such as games. In fact, a GA spends most of its time computing fitness. 


To mitigate this cost, fitness approximation techniques have been proposed to estimate the fitness values of individuals based on a set of features or characteristics. This paper focuses on performing fitness approximation in genetic algorithms using machine learning (ML) models.
Specifically, we propose to maintain a dataset of individuals and their actual fitness values, and to learn a surrogate fitness-approximation model based on this dataset.

Relying only on approximate fitness scores might cause the GA to converge to a false optimum. To address this, the evolutionary process can be controlled by combining approximate and actual fitness evaluations. This process is referred to as evolution control \citep{jin2000evolutionary}. While there are static evolution-control methods  \citep{ratle1998accelerating,bull1999model,ratle1999optimal}, our method \textit{dynamically} switches between true and approximate fitness evaluations, adapting to the evolutionary process's current state.

 We analyze several options for: 
1) switch conditions between using the actual fitness function and the approximate one,
2) sampling the search space for creating the dataset, and 
3) weighting the samples in the dataset. 

We evaluate our approach on three games implemented by Gymnasium, a framework designed for the development and comparison of reinforcement learning (RL) algorithms. We only use Gymnasium's game implementations, called environments, for evaluating fitness---we do not use the framework's learning algorithms.

\textbf{These are the main innovations of our proposed method}, which are further detailed in \autoref{sec:method}:
\begin{itemize}
    \item The method is generic and can be easily modified, extended, and applied to other domains. Other choices of ML model, switch condition, or sampling strategy can be readily made.
    \item The use of a switch condition (\autoref{sec:switch}) favors high flexibility in the experimental setting:
    \begin{itemize}
        \item Different switch conditions can be used according to the domain and complexity of the problem being solved.
        \item The predefined switch-threshold hyperparameter controls the desired amount of trade-off between result quality and computational time.
    \end{itemize}
    \item A monotonically increasing sample-weights function (\autoref{sec:sample-weights}) that places more emphasis on newer individuals in the learning process.
\end{itemize}

\textbf{Important points about our framework.} We wish to point out herein at the outset three key choices that we made in this research:
\begin{enumerate}
    \item \textit{Linear regression}. Regression is commonly used in fitness approximation \citep{tong2021surrogate}. We chose this simple model because it is very fast and as we shall see allows for retraining at will with virtually zero cost. Further, it proved quite potent. We plan to incorporate more models in the future.

    \item \textit{Benchmark problems}. It is quite simple nowadays to access numerous datasets. However, our goal was quality not quantity: ML datasets would usually not serve to showcase our approach, nor are they, in fact, the intended target problems, because fitness computation is cheap, or at least not costly enough. We specifically sought out simulation problems, where fitness computation is \textit{very} costly. This necessitated a hefty amount of behind-the-scenes exploration, testing, and coding, as such simulators are usually not written with GAs in mind.

    \item \textit{Baseline algorithms}. There are two approaches to performing fitness approximation: model-based approximation and similarity-based approximation \citep{tong2021surrogate}. As will be described below, our method is ML-based, yet takes solution similarity into account. We compared our method with other methods from both approximation approaches.
\end{enumerate}

The next section surveys relevant literature on fitness approximation. 
\autoref{sec:background} provides brief backgrounds on linear ML models and Gymnasium.
\autoref{sec:problems} introduces the problems being solved herein: Blackjack,  Frozen Lake, and Monster Cliff Walking.
\autoref{sec:method} describes the proposed framework in detail, followed by experimental results in \autoref{sec:experiments}.
\autoref{sec:extensions} presents two extensions to our method, involving novelty search and hidden fitness scores.
We end with concluding remarks and future work in \autoref{sec:conclude}.

\section{Fitness Approximation: Previous Work}
\label{sec:prev}
Fitness approximation is a technique used to estimate the fitness values of individuals without performing the computationally expensive fitness evaluation for each individual. By using fitness approximation, the computational cost of evaluating fitness throughout an evolutionary run can be significantly reduced, enabling the efficient exploration of large search spaces. 

There has been much interest in fitness approximation over the years, and providing a full review is beyond the scope herein. We focus on several works we found to be of particular relevance, some of which we compare to our proposed approach in \autoref{sec:experiments}.

\paragraph{Similarity-based fitness approximation}
\citet{smith1995fitness} suggested the use of fitness inheritance, where only part of the population has its fitness evaluated, and the rest inherit the fitness values from their parents. Their work proposed two fitness-inheritance methods:
1) averaged inheritance, wherein the fitness score of an offspring is the average of its parents;
and 2) proportional inheritance, wherein the fitness score of an offspring is a weighted average of its parents, based on the similarity of the offspring to each of its parents.
Although initially introduced in 1995, this method remains prevalent in recent academic publications \citep{giannakoglou2010multilevel,hameed2013large,wang2018fitness}.

\citet{kim2001efficient} presented fitness imitation, wherein the population is divided into several clusters. Only one individual in each cluster receives a true fitness score, and the rest of the individuals in the cluster receive an approximate fitness score based on a distance metric.


\citet{runarsson2004constrained} applied the K-nearest-neighbors algorithm to fitness approximation. Their technique approximates the fitness score of an individual based on its K nearest individuals with true fitness scores.

\paragraph{Fitness approximation using ML}
Surrogate models can be integrated into the evolutionary process in several ways: ML models can perform selection operators, approximate fitness evaluations, and genetic operators \citep{kneiding2024augmenting,rasheed2005methods}. We will focus on papers that use surrogate models for fitness evaluations. 

\citet{jin2005comprehensive} discussed various fitness-approximation methods involving ML models with offline and online learning, both of which are included in our approach.

\citet{bhattacharya2007surrogate} presented the Dynamic Approximate Fitness based Hybrid GA (DAFHEA), using Support Vector Machine as a surrogate model, and tested their approach on three benchmark functions and their noisy versions: Spherical, Rosenbrock, and Rastrigin. Their results showed satisfactory results both for noisy and non-noisy versions of the three functions.


\citet{dias2014genetic} used neural networks as surrogate models to solve a beam-angle optimization problem for cancer treatments. Their results were superior to an existing treatment type. They concluded that integrating surrogate models with genetic algorithms is an interesting research direction.

\citet{guo2017hybrid} proposed a hybrid GA with an Extreme Learning Machine (ELM) fitness approximation to solve the two-stage capacitated facility location problem. ELM is a fast, non-gradient-based, feed-forward neural network that contains one hidden layer, with random constant hidden-layer weights and analytically computed output-layer weights. The hybrid algorithm included offline learning for the initial population and online learning through sampling a portion of the population in each generation. This algorithm achieved adequate results in a reasonable runtime.

\citet{yu2018worth} examined the use of Support Vector Regression \citep{drucker1996support}, Deep Neural Networks \citep{prince2023understanding}, and Linear Regression models trained offline on sampled individuals, to approximate fitness scores in GAs. Specifically, the use of Linear Regression achieved adequate results when solving the One-Max and Deceptive problems.

\citet{zhang2022deep} used a deep surrogate neural network with online training to reduce the computational cost of the MAP-Elites (Multi-dimensional Archive of Phenotypic Elites) algorithm for constructing a diverse set of high-quality card decks in Hearthstone. Their work achieved state-of-the-art results.

\citet{livne2022evolving} compared two approaches for fitness approximation, because a full approximation in their case would require 50,000 training processes of a deep contextual model, each taking about 1 minute: 1) training a multi-layer perception sub-network instead, which takes approximately five seconds; 2) a pre-processing step involving the training of a robust single model. The latter improved training time from 1 minute to 60 milliseconds.

\section{Preliminaries}
\label{sec:background}

\subsection{Linear ML}
Linear ML models are a class of algorithms that learn a linear relationship between the input features and the target variable(s). We focus on two specific linear models, namely Ridge (also called Tikhonov) regression \citep{hoerl1970ridge} and Lasso regression (least absolute shrinkage and selection operator) \citep{tibshirani1996regression}. These two models strike a balance between complexity and accuracy, enabling efficient estimation of fitness values for individuals in the GA population. 

Ridge and Lasso are linear regression algorithms with an added regularization term to prevent overfitting. Their loss functions are given by:
\[
L_1: ||y - Xw||^2_2 + \alpha * ||w||_1\,,
\]
\[
L_2: \|y - Xw\|_2^2 + \alpha \|w\|_2^2\,,
\]
where $L_1$ is for Lasso, $L_2$ is for Ridge, $X$ represents the feature matrix, $y$ represents the target variable, $w$ represents the coefficient vector, and $\alpha$ represents the regularization parameter. 




Regression is commonly used as a surrogate model for fitness evaluations \citep{tong2021surrogate}. A major advantage of linear models with respect to our framework is that they are super-fast, enabling us to treat model-training time as virtually zero (with respect to fitness-computation time in the simulator). Thus, we could retrain a model as often as we chose.
As recently noted by \citet{ISLP2023}: ``Historically, most methods for estimating \textit{f} have taken a linear form. In some situations, such an assumption is reasonable or even desirable.''
It is worth mentioning that our generic method can easily be integrated with other types of ML models.

\subsection{Gymnasium}
Gymnasium (formerly OpenAI Gym) \citep{brockman2016openai} is a framework designed for the development and comparison of reinforcement learning (RL) algorithms. It offers a variety of simulated environments that can be utilized to evaluate the performance of AI agents. Gymnasium offers a plethora of simulators, called \href{https://gymnasium.farama.org/api/env/}{environments}, from different domains, including robotics, games, cars, and more. Each environment defines state representations, available actions, observations, and how to obtain rewards during gameplay.

A Gymnasium simulator can be used for training an RL agent or as a standalone simulator.
Herein, we take the latter approach, using these simulators to test our novel fitness-approximation method for an evolutionary agent system.

\section{Problems} 
\label{sec:problems}
\begin{figure}
    \centering
    \subfloat[\centering Blackjack]{{\includegraphics[scale=0.235]{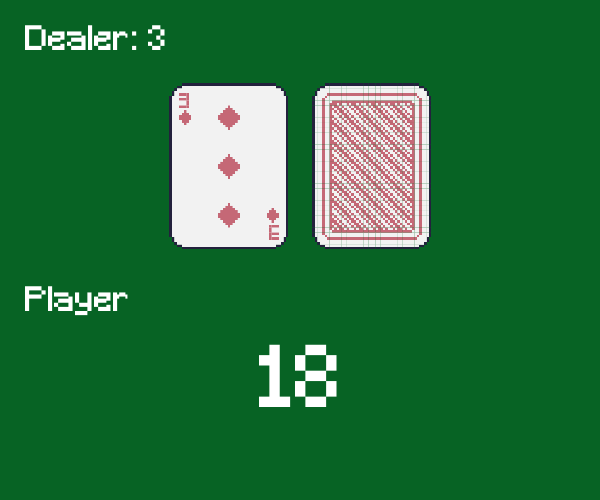} }}
    \qquad
    \subfloat[\centering Frozen Lake]{{\includegraphics[scale=0.23]{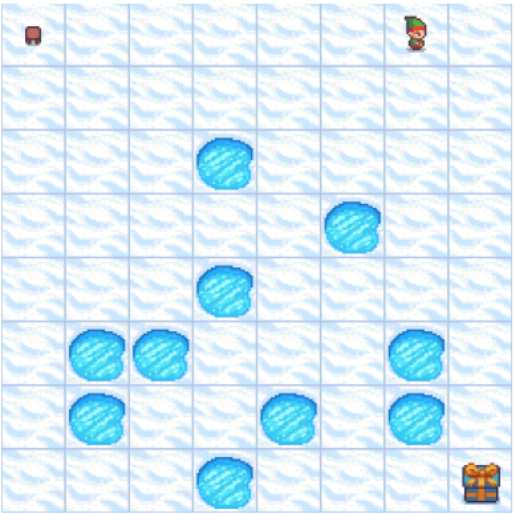} }}
    \qquad
    \subfloat[\centering Monster Cliff Walking]{{\includegraphics[scale=0.485]{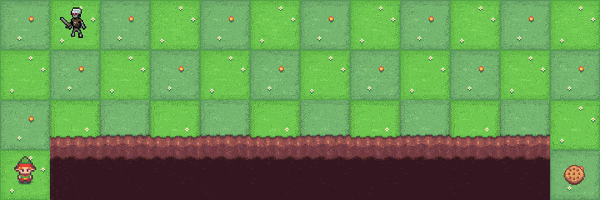} }}
    \caption{Gymnasium environments (or custom modifications of them) we use for actual fitness-score evaluation.}
    \label{fig:problems}
\end{figure}

This section provides details on the three problems from Gymnasium that we will tackle: Blackjack, Frozen Lake, and Monster Cliff Walking (\autoref{fig:problems}). 
As noted, we specifically sought out simulation problems---where fitness computation is \textit{very} costly; this required some lengthy behind-the-scenes exploration, testing, and coding, as such simulators are usually not written with GAs in mind.

\subsection{Blackjack}
\label{sec:blackjack}
Blackjack is a popular single-player card game played between a player and a dealer. The objective is to obtain a hand value closer to 21 than the dealer's hand value---without exceeding 21 (going bust). We follow the game rules defined by \citet{sutton2018reinforcement}. Each face card counts as 10, and an ace can be counted as either 1 or 11. The \href{https://gymnasium.farama.org/environments/toy_text/blackjack/}{Blackjack environment} of Gymnasium represents a state based on three factors: 
1) the sum of the player's card values,
2) the value of the dealer's face-up card, and 
3) whether the player holds a usable ace. An ace is usable if it can count as 11 points without going bust.
Each state allows two possible actions: stand (refrain from drawing another card) or hit (draw a card).

We represent an individual as a binary vector, where each cell corresponds to a game state from which an action can be taken; the cell value indicates the action taken when in that state. As explained by \citet{sutton2018reinforcement}, there are 200 such states, therefore the size of the search space is $2^{200}$. 

The actual fitness score of an individual is computed by running 100,000 games in the simulator (the same number of games as in the \href{https://gymnasium.farama.org/tutorials/training_agents/blackjack_tutorial/}{Gymnasium Blackjack Tutorial}) and then calculating the difference between the number of wins and losses. We normalize fitness by dividing this difference by the total number of games. The ML models and the GA receive the normalized results (i.e., scores $\in [-1,1]$), but we will display the non-normalized fitness scores for easier readability. Given the inherent advantage of the dealer in the game, it is expected that the fitness scores will mostly be negative.

\subsection{Frozen Lake}
\label{sec:frozen-lake}
In this game, a player starts at the top-left corner of a square board and must reach the bottom-right corner. Some board tiles are holes. Falling into a hole leads to a loss, and reaching the goal leads to a win. Each tile that is not a hole is referred to as a frozen tile.

Due to the slippery characteristics exhibited by the frozen lake, the agent might move in a perpendicular direction to the intended direction. For instance, suppose the agent attempts to move right, after which the agent has an equal probability of $\frac{1}{3}$ to move either right, up, or down. This adds a stochastic element to the environment and introduces a dynamic element to the agent's navigation.

For consistency and comparison, all simulations will run on the 8x8 map presented in \autoref{fig:problems}. In this map, the \href{https://gymnasium.farama.org/environments/toy_text/frozen_lake/}{Frozen Lake environment} represents a state as a number between $0$ and $63$. There are four possible actions in each state: move left, move right, move up, or move down. Our GA thus represents a Frozen Lake agent as an integer vector with a cell for each frozen tile on the map, except for the end-goal state (since no action can be taken from that state). Similarly to Blackjack, each cell dictates the action being taken when in that state. Since there are 53 frozen tiles excluding the end-goal, the size of the search space is $4^{53}=2^{106}$. The fitness function is defined as the percentage of wins out of 2000 simulated games (the same number of games as in the \href{https://gymnasium.farama.org/tutorials/training_agents/FrozenLake_tuto/}{Gymnasium Frozen Lake Tutorial}). Again, we will list non-normalized fitness scores.

\subsection{Monster Cliff Walking}
\label{sec:monster-cliff-walking}
In \href{https://gymnasium.farama.org/environments/toy_text/cliff_walking/}{Cliff Walking}, the player starts at the bottom-left corner of a 4x12 board and must reach the bottom-right corner. All the tiles in the bottom row that are not the starting position or goal are considered cliffs. The player must reach the goal without falling into the cliff.

Since this game can be solved quickly by a GA, we tested a stochastic, more-complex version of the game, called \href{https://github.com/Sebastian-Griesbach/MonsterCliffWalking}{Monster Cliff Walking}. In this version, a monster spawns in a random location and moves randomly among the top three rows of the board. Encountering the monster leads to an immediate loss.

The player performs actions by moving either up, right, left, or down. A state in this environment is composed  both of the player's location and the monster's location.

There are 37 tiles where an action can be taken by the player (excluding the cliff and end goal) and 36 possible locations for the monster. Therefore, there are 1332 different states in the game. Similarly to Frozen Lake, an agent in Monster Cliff Walking is represented as an integer vector whose size is equal to the number of the states in the game. The size of the search space is $4^{1332}=2^{2664}$, significantly larger than the search spaces of the previous two problems.

Due to stochasticity, each simulation runs for 1000 episodes. An episode ends when one of the following happens: 
1) The player reaches the end-goal state; 
2) the player encounters the monster; 
3) the player performs 1000 steps (we limited the number of steps to avoid infinitely cyclic player routes). Falling into a cliff does not end the episode, it only restarts the player's position.

The fitness function is defined as the average score of all episodes. When an episode ends, the score for the episode is computed as the total rewards obtained during the episode. A penalty of -1 is obtained per step, -100 is added every time the agent falls into a cliff, and -50 is added if the player has encountered the monster (-1000 in the original environment, but we used reward shaping to allow easier exploration of the search space). Due to the reward mechanism and fitness-function definition, it is expected that the fitness scores will be negative.

Given the variety of the fitness-function values, the ML model is trained on the natural logarithm of fitness scores, and its predictions are raised to an exponent.
More formally:
\[
y_{train} = log(-f_{true}),
\]
\[
f_{approx} = -e^{y_{pred}},
\]
where $y_{train}$ is the regression target-value vector in the ML model training,
$f_{true}$ is the true fitness score sent to the model,
$f_{approx}$ is the approximate fitness score sent to the evolutionary algorithm,
and $y_{pred}$ is a vector of predictions returned by the model.

\section{Proposed Method}
\label{sec:method}
This section presents our proposed method for fitness approximation in GAs using ML models. We outline the steps involved in integrating Ridge and Lasso regressors into the GA framework, and end with a discussion of advantages and limitations of the new method.

\subsection{Population dataset}
\label{sec:dataset}
Our approach combines both offline and online learning, as depicted in \autoref{fig:fit-approx}. The algorithm begins in \textit{evolution mode}, functioning as a regular GA, where a population evolves over successive generations. However, each time a fitness score is computed for an individual, we update a dataset whose features are the encoding vector of the individual and whose target value is the respective fitness score. An illustration of the dataset is presented in \autoref{tab:dataset}. The initial population will always be evaluated using the simulator since the population dataset is empty at this stage of the run.

\tikzset{
    treenode/.style = {shape=rectangle, rounded corners,
                     draw, anchor=center,
                     text width=4em, minimum height=3em, align=center,
                     top color=white, bottom color=blue!20,
                     inner sep=1ex},
    decision/.style = {treenode, diamond, inner sep=0pt, bottom color=yellow!40,},
    root/.style     = {treenode, font=\Large, bottom color=red!30},
    env/.style      = {treenode, font=\ttfamily\normalsize},
    finish/.style   = {root, bottom color=green!40},
    dummy/.style    = {circle,draw},
    database/.style={
      cylinder, cylinder uses custom fill, cylinder body fill=purple!35, cylinder end fill=purple!15, shape border rotate=90, aspect=0.25, draw, text width=3em, minimum height=3em,
    },
    >={Stealth[length=3mm]},
    pics/population/.style 2 args={code={
        \def\n{#1} 
        \def\name{pop-#2}
        \def\a{1} 
        \def\b{0.5} 
        \def\circleradius{0.09} 
        \pgfmathsetseed{18}
        
        \node[draw, ellipse, anchor=center, minimum width=2*\a cm, minimum height=2*\b cm] (\name) {};

        \foreach \i in {1,...,\n} {
          \pgfmathsetmacro{\randangle}{\i*360/\n}
          \pgfmathsetmacro{\radius}{\a*\b/sqrt(\a^2*sin(\randangle)^2 + \b^2*cos(\randangle)^2)}
          \pgfmathsetmacro{\randradius}{min(\radius-\circleradius-0.03,rnd*\radius+0.1)}
          \pgfmathsetmacro{\randx}{\randradius*cos(\randangle)}
          \pgfmathsetmacro{\randy}{\randradius*sin(\randangle)}

          \draw[fill=white] (\randx, \randy) circle (\circleradius);
        }
    }},
}

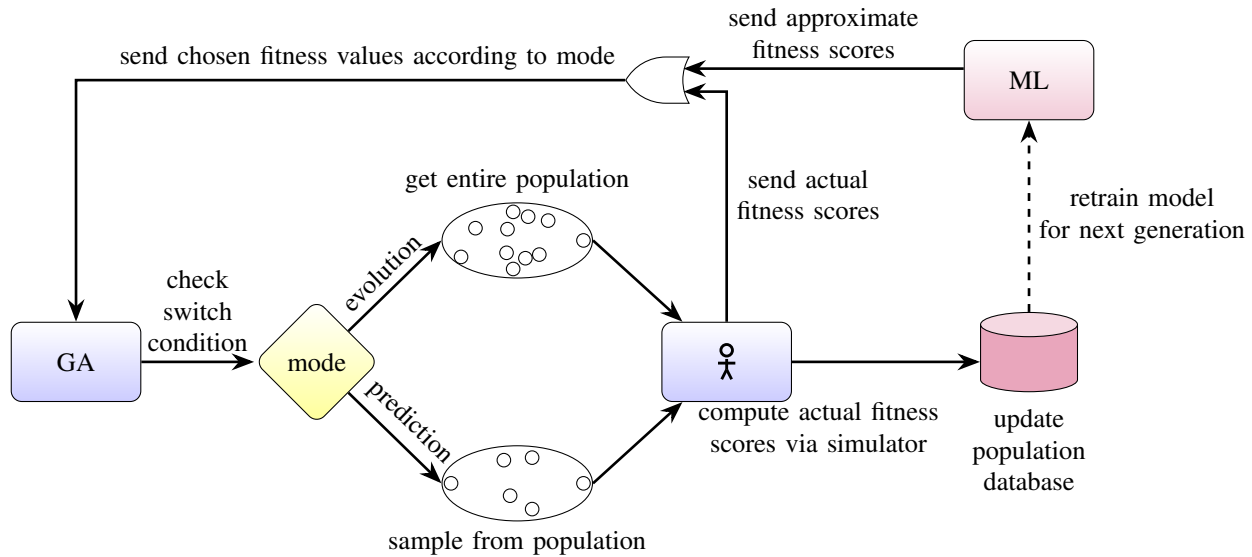
\begin{figure*}
\centering
\def\d{1.5}
\begin{tikzpicture}[node distance=\d cm,circuit logic US]
    \node[treenode] (ga) {GA};
    \node[decision, right=of ga] (mode) {mode};
    \pic[above right=0.8*\d cm and 1.5*\d cm of mode]{population={11}{entire}};
    \node[above] at (pop-entire.north) {get entire population};
    \pic[below right=0.8*\d cm and 1.5*\d cm of mode]{population={6}{sampled}};
    \node[below] at (pop-sampled.south) {sample from population};
    \node[treenode, right=2.5*\d cm of mode,label={[align=center,yshift=-1.4cm,xshift=-0.5cm,anchor=west]:compute actual fitness\\scores via simulator}] (gym) {\Strichmaxerl[2][-180][180]};
    \node[database, right=2.5cm of gym, label={[align=center,yshift=-2.5cm]:update\\population\\database}] (db){};
    \node[treenode, above=1.7*\d cm of db, bottom color=purple!20] (ml) {ML};
    \node[or gate, left=3*\d cm of ml, rotate=180] (or) {};
    \path[->,line width=1px]
        (ga) edge node[align=center,above]{check\\switch\\condition} (mode)
        (mode) edge node[align=center,above,sloped] {prediction} (pop-sampled.west)
        (mode) edge node[align=center,above,sloped] {evolution} (pop-entire.west)
        (pop-sampled.east) edge (gym)
        (pop-entire.east) edge (gym)
        (gym) edge (db)
        (db) edge[dashed] node[align=center,right]{retrain model\\for next generation}(ml)
        (ml.170) edge node[align=center,above]{send approximate\\fitness scores} (ml.170 -| or.west);
    \draw[->,line width=1px] (or.east) -| node[align=center,above,pos=0,anchor=south east]{send chosen fitness values according to mode} (ga);
    \draw[->,line width=1px] (gym) |- node[align=center,right,anchor=south west,pos=0.2]{send actual\\fitness scores} (ml.190 -| or.west);
\end{tikzpicture}
\caption{Flowchart of proposed method.
In evolution mode the algorithm functions as a regular GA. 
When the switch condition is met the algorithm shifts to prediction mode: actual (in-simulator) fitness values are calculated only for a sampled subset of the population, while the rest are assigned approximate fitnesses from the ML model. This latter is retrained before moving to the next generation.}
\label{fig:fit-approx}
\end{figure*}

\begin{table}
\centering
  \caption{An example of a dataset created during a GA run of the Frozen Lake environment.
  An individual is a vector of 53 attributes. 
  A total of $m$ fitness computations have been carried out so far.}
  \label{tab:dataset}
  \begin{tabular}{cccccc}
    \toprule
    \textbf{Index} & $\pmb{a_1}$ & $\pmb{a_2}$ & \textbf{\dots} & $\pmb{a_{53}}$ & \textbf{Fitness} \\
    \midrule
    1 & 0 & 3 & \dots & 1 & 0.003 \\
    \midrule
    2 & 0 & 1 & \dots & 3 & 0.001 \\
    \midrule
    \dots & \dots & \dots & \dots & \dots & \dots \\
    \midrule
    m & 3 & 1 & \dots & 1 & 0.545 \\
    \bottomrule
  \end{tabular}
\end{table}

After a predefined \textit{switch condition} is met the algorithm transitions from \textit{evolution mode} to \textit{prediction mode}. In prediction mode, actual (in-simulator) fitness scores are computed only for a sampled subset of the population. For the rest of the population, the GA assigns approximate fitness values using a learned ML model that was trained on the existing population dataset. In prediction mode, a sample of the population receives true fitness scores per generation, and the rest of the generation's individuals receive approximate fitness scores. This method is referred to as individual-based controlled evolution \citep{jin2000evolutionary}. The algorithm can switch back and forth between evolution mode and prediction mode, enabling dynamic adaptation to the evolutionary state.



Witness the interplay between the dynamic \textit{switch condition} and the static (pre-determined) \textit{sample rate}---a hyperparameter denoting the percentage of the population being sampled. In cases where lower runtimes are preferred, using a relatively lenient switch condition is better, resulting in a higher fitness-approximation rate coupled with reduced runtime---at some cost to fitness quality. On the contrary, in cases where accurate fitness scores are preferred, the use of a strict switch condition is advisable, to allow ML fitness approximation only when model confidence is high.

Note that the number of actual, in-simulator fitness computations performed is ultimately determined dynamically by the coaction of \textit{switch condition} and \textit{sample rate}.

In stochastic domains such as ours, the same individual may receive different (actual) fitness scores for every evaluation, and thus appear in the population dataset multiple times---with different target values. This can be prevented by keeping a single copy of each individual in the dataset, or by computing an average fitness score of all evaluations of the same individual (or possibly some other aggregate measure). However, since these solutions greatly interfere with the sample weighting mechanism (described in \autoref{sec:sample-weights}), we decided to remove identical duplicate rows only (i.e., with both equal representations and fitness scores) while keeping individuals with equal representation and different fitness scores in the dataset.


\subsection{Switch condition}
\label{sec:switch}
The switch condition plays a crucial role in determining when the GA transitions from evolution mode to prediction mode (and vice-versa). Our approach defines the switch condition based on a predefined criterion. Once this criterion is met, the algorithm switches its focus from evolving based entirely on full fitness computation to obtaining approximate fitness values through the model.

The switch condition can be defined in various ways depending on the specific problem and requirements. It may involve measuring the accuracy of the model's predictions, considering a predefined threshold, or other criteria related to the state of the population and the model. 
In situations where the model's accuracy falls below the desired threshold, the algorithm can revert back to evolution mode until the condition for switching to prediction mode is met once again.

Determining an appropriate switch condition is crucial for balancing the trade-off between the accuracy of fitness approximation and the computational efficiency of the algorithm. It requires tuning to find the optimal configuration for a given problem domain. Overall, the switch condition serves as a pivotal component in our approach, enabling a smooth transition between evolution mode and prediction mode based on a predefined criterion. 

We defined and tested four different switch conditions, each having a hyperparameter called \textit{switch\_threshold}:

\begin{enumerate}
\item \textbf{Dataset size.} The simplest solution entailed performing regular evolution until the dataset reaches a certain size threshold, and then transitioning to prediction mode indefinitely. Although simple, this switch condition is less likely to adapt to the evolutionary state due to its inability to switch back to evolution mode.

\item \textbf{Plateau.} Wait for the best fitness score to stabilize before transitioning to prediction mode. We consider the best fitness score as stable if it has not changed much (below a given threshold) over the last $P$ generations, for a given $P$.
This method posed a problem, as the model tend to maintain the evolutionary state without significant improvement throughout the run.

\item \textbf{CV error.} Evaluate the model's error using cross-validation on the dataset. We switch to predict mode when the error falls below a predetermined threshold, and vice versa. We will demonstrate the use of this switch condition in the Frozen Lake scenario.

\item \textbf{Cosine similarity.} Cosine similarity is a metric commonly used in Natural-Language Processing to compare different vectors representing distinct words \citep{salton1975vector}. We use this metric to compare the vectors in the GA population with those in the ML dataset. The underlying idea is that the model will yield accurate predictions if the current population closely resembles the previous populations encountered by the model, up to a predefined threshold. Our method utilizes this switch condition in the Blackjack and Monster Cliff Walking scenarios.
\end{enumerate}

The switch condition is a key component of our method. This section presented the switch conditions tested during our research, yet our method can easily be enhanced with other domain-specific switch conditions.

\subsection{Sampling strategy}
\label{sec:sample-strategy}
Training the ML model on the initial population only is insufficient for performing high-quality approximations: the model needs to be updated throughout the evolutionary process. This can be done by selecting sample individuals and computing their true fitness score during different stages of the evolutionary run (individual-based evolution control) \citep{jin2000evolutionary}.

As mentioned in \autoref{sec:dataset}, during prediction mode, a subset of the population is sampled in each generation. There are several sampling strategies, and choosing the right strategy can greatly impact the quality of the population dataset. We focus on three strategies we tested within our method:

\begin{enumerate}
    \item \textbf{Random sampling.} The most straightforward sampling strategy is to randomly pick a subset of the population and compute their actual fitness scores while approximating the fitness scores for the rest of the population. Despite its simplicity, this strategy does not leverage any information about the dataset, the population, or the domain.

    \item \textbf{Best strategy.} This approach samples the individuals with the best approximated fitness score \citep{jin2000evolutionary}. This strategy was used in the work of \citet{guo2017hybrid}, one of the papers we compared our results to.

    \item \textbf{Similarity sampling.} Another approach is choosing the individuals that are the least similar to the individuals that already exist in the dataset. Using this method will improve the diversity of the dataset and hence improve the ability of the model to generalize better to a wider volume of the search space. This strategy is useful for domains where individuals with similar representations receive similar fitness scores, such as our domains. The similarity metric we chose is the cosine similarity, discussed above.
\end{enumerate}

Our genetic method allows for the seamless integration of additional strategies, such as clustering strategy \citep{graning2007individual}, and others \citep{ratle1999optimal}.


\subsection{Sample weights} 
\label{sec:sample-weights}
In this section, `sample' refers to a row in the dataset (as is customary in ML)---not to be confused with `sampling' in `sampling strategy', introduced in the previous section. 

During the training of the ML model on the dataset, each individual typically contributes equally. However, individuals tend to change and improve over the course of evolution. We preferred that the model pay more attention to recent generations, since recently added individuals are more relevant to the current state of the evolutionary process, compared to the individuals at the beginning. To account for this, we track the generation in which each individual is added to the dataset and assign weights accordingly. The weight assigned to an individual increases with the generation number. Note that individuals from earlier generations are not ignored by the model and are not removed from the dataset---they only have a smaller impact on the learning process of the model. After experimenting with various weighting functions, we established a square root relationship between the generation number and its corresponding weight: $\mathit{weight} = \mathit{\sqrt{gen}}$. We note that the algorithms we used do not require the weights to sum to one. Sample weights were also applied by \citet{jin2001managing}.

\subsection{Advantages and limitations}
\label{sec:advantages-limitations}
Our proposed method offers several advantages. It can potentially reduce the computational cost associated with evaluating fitness scores in a significant manner. 
Rather than compute each individual's fitness every generation, the population receives an approximation from the ML model at a negligible cost.

The use of models like Ridge and Lasso helps avoid overfitting by incorporating regularization. This improves the generalization capability of the fitness-approximation model.

Additionally, our approach allows for continuous learning, by updating the dataset and retraining the model during prediction mode. The continual retraining is possible because the ML algorithms are extremely rapid and the dataset is fairly small.

There are some limitations to consider. Linear models assume a linear relationship between the input features and the target variable. Therefore, if the fitness landscape exhibits non-linear behavior, the model may not capture it accurately. In such cases, alternative models capable of capturing non-linear relationships may be more appropriate; we plan to consider such models in the future.

Further, the performance of the fitness-approximation model heavily relies on the quality and representativeness of the training dataset. If the dataset does not cover the entire search space adequately, the model's predictions may be less accurate. Careful consideration should be given to dataset construction and sampling strategies to mitigate this limitation. We took this into account when choosing the appropriate switch conditions and sampling strategy, discussed above.

An additional limitation is the choice of the best individual to be returned at the end of the run. Since a portion of the fitness values is approximate, the algorithm might return an individual with a good predicted fitness score, but with a bad actual fitness score. To address this issue we return the individual with the best fitness from the population dataset (which always holds actual fitness values).

\section{Experiments and Results}
\label{sec:experiments}
To assess the efficacy of the proposed approach, we carried out a comprehensive set of experiments aimed at solving the three problems outlined in \autoref{sec:problems}. Our objective was to compare the performance of our method against a ``full'' GA (computing all fitnesses), considering solution quality and computational efficiency as evaluation criteria. 

Experiments were conducted using the EC-KitY software \citep{eckity2023} on a cluster of 96 nodes and 5408 CPUs (the most powerful processors are AMD EPYC 7702P 64-core, although most have lesser specs). 10 CPUs and 3 GB RAM were allocated for each run. Since the nodes in the cluster vary in their computational power, and we had no control over the specific node allocated per run, we measured computational cost as number of actual (in-simulator) fitness computations performed, excluding the initial-population fitness computation. The average duration of a single fitness computation was 21 seconds for Blackjack, 6 seconds for Frozen Lake, and 9 seconds for Monster Cliff Walking.
The source code for our method and experiments can be found at \url{https://github.com/Anon..}

Both fitness-approximation runs and full-GA runs included the same genetic operators with the same probabilities: tournament selection \citep{blickle2000tournament}, two-point crossover \citep{spears1991study}, bit-flip mutation for Blackjack and uniform mutation for Frozen Lake \citep{lim2017crossover}. The specific linear models utilized in the experiments and their hyperparameters  are detailed in \autoref{tab:hyperparameters}. Their values were chosen by extracting datasets from traditional GA runs (such as the one described in \autoref{tab:dataset}), and using k-fold cross-validation in an offline manner, similarly to what was done by \citet{yu2018worth}. We used the linear models provided by scikit-learn \citep{pedregosa2011scikit}. Hyperparameter tuning was done using Optuna \citep{akiba2019optuna}.

\begin{table*}
\centering
\caption{Hyperparameters.}
\label{tab:hyperparameters}
\begin{tabular}{rlccc}
Hyperparameter & Explanation & Blackjack & Frozen Lake & Monster Cliff Walking \\ \hline
\textit{switch\_condition} & see \autoref{sec:switch} & Cosine & CV Error & Cosine \\
\textit{switch\_threshold} & see \autoref{sec:switch} & 0.9 & 0.02 & 0.96 \\
\textit{sample\_strategy} & see \autoref{sec:sample-strategy} & Similarity & Similarity & Similarity \\
\textit{population\_size} & number of individuals in population & 100 & 100 & 100 \\
\textit{p\_crossover} & crossover rate between two individuals & 0.7 & 0.7 & 0.7\\
\textit{p\_mutation} & mutation probability per individual & 0.3 & 0.3 & 0.3\\
\textit{generations} & number of generations & 200 & 50 & 100\\
\textit{tournament\_size} & for tournament selection & 4 & 4 & 4\\
\textit{crossover\_points} & number of crossover points for $K$-point crossover & 2 & 2 & 2\\
\textit{mutation\_points} & number of mutation points for $N$-point mutation & 20 & 6 & 133 \\
\textit{model} & type of model that predicts fitness scores & Ridge & Lasso & Ridge \\
\textit{alpha} & model regularization parameter & 0.3 & 0.65 & 0.6 \\
\textit{max\_iter} & maximum iterations for ML training algorithm & 3000 & 1000 & 2000 \\
\textit{gen\_weight} & weighting function, discussed in \autoref{sec:sample-weights} & Square root & Square root & Square root \\
\end{tabular}
\end{table*}

Note the difference between the hyperparameter \textit{generations}, which designates the number of GA generations, and \textit{max\_iter}, which designates the number of iterations in the ML model training. The latter is relatively negligible in terms of runtime differences, due to the short training time of the linear ML models.

We performed 20 replicates per sample rate, and assessed statistical significance by running a 10,000-round permutation test, comparing the mean scores between our proposed method and the full GA (with full fitness computation). The results are shown in \autoref{tab:perm-tests}. 

\begin{table}
\centering
\caption{Results. 
Each row summarizes 20 replicate runs. The last row represents the full GA.
\textbf{Sample rate}: proportion of fitness values computed in prediction mode. 
\textbf{Absolute fitness}: mean best-of-run fitness values of all replicates.
\textbf{Relative fitness}: Absolute fitness compared to GA (percentage).
\textbf{Absolute computations}: mean number of actual fitness computations done in simulator across all replicates.
\textbf{Relative computations}: Absolute computations compared to GA (percentage).
\textbf{p-value}: result of 10,000-round permutation test, comparing the mean scores between our proposed method and the full GA (with full fitness computation).
Boldfaced results are those that are statistically identical (meaning statistical insignificance) in performance to full GA, i.e., p-value $> 0.05$.}
\label{tab:perm-tests}
\footnotesize  
\subfloat[\centering Blackjack]{\begin{tabular}{cccccc}
Sample & Absolute & Relative & Absolute & Relative & p-value \\ rate & fitness & fitness & computations & computations\\
\hline
20\% & -4517.45 & 84.92\% & 4196 & 20.98\% & 1e-4 \\
40\% & -4018.9 & 95.46\% & 8075 & 40.38\% & 4.5e-3 \\
\textbf{60\%} & \textbf{-3904.65} & \textbf{98.25\%} & \textbf{12,022} & \textbf{60.11\%} & \textbf{0.13} \\
\textbf{80\%} & \textbf{-3803.55} & \textbf{100.86\%} & \textbf{16,006} & \textbf{80.03\%} & \textbf{0.46} \\
GA & -3836.4 & 100\% & 20,000 & 100\% \\
\end{tabular}}
    \qquad
    \subfloat[\centering Frozen Lake]{\begin{tabular}{cccccc}
Sample & Absolute & Relative & Absolute & Relative & p-value \\ rate & fitness & fitness & computations & computations\\
\hline
20\% & 1064.6 & 84.91\% & 2512 & 50.24\% & 1e-4  \\
\textbf{40\%} & \textbf{1223.2} & \textbf{97.56\%} & \textbf{4325} & \textbf{86.5\%} & \textbf{0.11}  \\
\textbf{60\%} & \textbf{1248.9} & \textbf{99.61\%} & \textbf{4730} & \textbf{94.6\%} & \textbf{0.8} \\
\textbf{80\%} & \textbf{1262.95} & \textbf{100.73\%} & \textbf{4892} & \textbf{97.84\%} & \textbf{0.58} \\
GA & 1253.75 & 100\% & 5000 & 100\%  \\
\end{tabular}}
\qquad
    \subfloat[\centering Monster Cliff Walking]{\begin{tabular}{cccccc}
Sample & Absolute & Relative & Absolute & Relative & p-value \\ rate & fitness & fitness & computations & computations\\
\hline
{20\%} & {-130.51} & {82.39\%} & {4228} & {42.28\%} & {1e-4}  \\
{40\%} & {-127.61} & {84.26\%} & {4735} & {47.35\%} & {1e-4}  \\
{60\%} & {-115.55} & {93.06\%} & {6336} & {63.36\%} & {0.02} \\
\textbf{80\%} & \textbf{-110.41} & \textbf{97.39\%} & \textbf{8170} & \textbf{81.7\%} & \textbf{0.24} \\
GA & -107.53 & 100\% & 10,000 & 100\%  \\
\end{tabular}}
\normalsize
\end{table}

Examining the results reveals an observable rise in fitness scores, along with an increase in the number of fitness computations, as sample rate increases. This is in line with the inherent trade-off within our method, wherein the quality of the results and the runtime of the algorithm are interconnected. Further, there is a strong correlation between the sample rate and the relative number of fitness computations. Notably, as the relative fitness-score computation approaches the sample rate, the frequency of individuals with approximate fitness scores increases. 

In the Blackjack scenario, fitness-computation ratios closely approximate the sample rates, indicating a strong dominance of prediction mode. In contrast, computation ratios for Frozen Lake are relatively close to 100\%, with the exception of the 20\% sample rate, signifying a prevalence of evolution mode in the majority of generations.
In Monster Cliff Walking there is a strong dominance of prediction mode, except for the 20\% sample rate.

These observations shed light on the impact of the switch condition and its predefined threshold hyperparameters on the behavior of the algorithm in approximating fitness scores.

Boldfaced results in \autoref{tab:perm-tests} are those that are statistically identical (meaning statistical insignificance) in performance to the full GA, i.e., \textit{p-value} $>0.05$. 

\textit{We observe that results indistinguishable from the full GA can be obtained with a significant reduction in fitness computation.}

\autoref{tab:comparison} shows the performance on our three benchmark problems of three methods discussed in \autoref{sec:prev}:
HEA/FA \citep{guo2017hybrid},
Averaged Fitness Inheritance \citep{smith1995fitness},
and Proportional Fitness Inheritance \citep{smith1995fitness}.
The first method performs fitness approximation with an ELM as a surrogate model and the other two perform similarity-based fitness approximation. 

Of the model-based approximation methods presented in \autoref{sec:prev}, HEA/FA was the most-similar method to ours. The fitness-inheritance methods use our genetic operators (discussed above), and HEA/FA uses different genetic operators, delineated by \citet{guo2017hybrid}.

We also tested the performance of ELM as the ML model with our genetic operators, no sample weights, and random sampling.
Note that these four methods do not use a dynamic switch condition, instead utilizing fitness approximation for the entire GA run.

Choosing the appropriate competitors for evaluation proved difficult:
Some papers included only a pseudocode that is complicated to implement (e.g., \citep{dias2014genetic,bhattacharya2007surrogate}), and some papers contained code examples that included major differences in runtime environments---such as GPU usage, different programming languages, and complex simulator integrations (e.g., \citep{livne2022evolving,zhang2022deep}).
In papers that did not include code implementation, we contacted the authors for the implementation of the papers (unfortunately, they are not available on GitHub) but received no reply---so we implemented the methods ourselves.

The HEA/FA algorithm produced unsatisfactory results. We assume it is due to possible differences between our implementation and the original one, such as the algorithm implementation, the genetic operators, or the hidden-layer size and activation function (these were not given in the paper, and we set them to 100 and ReLU, respectively).

ELM produced better results, particularly in Frozen Lake, but it did not achieve statistical insignificance for any problem. Although our approach uses a different ML model, this emphasizes the importance of dynamic fitness approximation, as implemented in our architecture.

Prop-FI, and Avg-FI performed relatively well, especially in Frozen Lake and Monster Cliff Walking. They achieved statistical insignificance at an 80\% sample rate with a lower computational cost compared to our approach.
However, statistical insignificance (i.e., same as full GA) was only attained at an 80\% sample rate in Frozen Lake and Monster Cliff Walking (and not at all in Blackjack)---whereas our method achieved statistical insignificance for all problems, partially for lower sample rates as well, as seen in \autoref{tab:perm-tests}.

In summary: Compare the boldfaced lines (or lack thereof) of \autoref{tab:perm-tests} and \autoref{tab:comparison}.

\begin{table*}
\centering
\caption{Comparison with previous work.
\textbf{HEA/FA}: Our implementation of the method suggested by \citet{guo2017hybrid}.
\textbf{Avg-FI}: Our implementation of Average Fitness Inheritance, as suggested by \citet{smith1995fitness}.
\textbf{Prop-FI}: Our implementation of Proportional Fitness Inheritance as suggested by \citet{smith1995fitness}, using Cosine Similarity as similarity metric.
\textbf{ELM}: ELM-based fitness approximation.
\textbf{ApproxML}: Best results for our algorithm.
Boldfaced results are those that are statistically identical (meaning statistical insignificance) in performance to full GA (last row), i.e., p-value $> 0.05$.}
\label{tab:comparison}
\resizebox{0.48\textwidth}{!}{
\subfloat[\centering Blackjack]{\begin{tabular}{ccccccc}
          & Sample & Absolute & Relative & Absolute & Relative &  \\
  Method  & rate & fitness & fitness & computations & computations & p-value \\
\hline
HEA/FA & 20\% & -11,330.8 & 33.86\% & 6111.05 & 30.56\% & 1e-4 \\
HEA/FA & 40\% & -10,423.58 & 36.81\% & 10,203.26 & 51.02\% & 1e-4 \\
HEA/FA & {60\%} & {-9409.45} & {40.77\%} & {14,277.65} & {71.39\%} & 1e-4 \\
HEA/FA & {80\%} & {-9250.42} & {41.47\%} & {18,364.11} & {91.82\%} & 1e-4 \\
\hline
Avg-FI & 20\% & -8523.8 & 45.01\% & 4000 & 20\% & 1e-4 \\\
Avg-FI & 40\% & -7508.95 & 51.09\% & 8000 & 40\% & 1e-4 \\
Avg-FI & {60\%} & {-5601.5} & {68.49\%} & {12,000} & {60\%} & 1e-4 \\
Avg-FI & {80\%} & {-4031.05} & {95.17\%} & {16,000} & 80\% & 2e-4 \\
\hline
Prop-FI & 20\% & -9093.45 & 42.19\% & 4000 & 20\% & 1e-4 \\
Prop-FI & 40\% & -7206.45 & 53.24\% & 8000 & 40\% & 1e-4 \\
Prop-FI & {60\%} & {-5738.75} & {66.85\%} & {12,000} & {60\%} & 1e-4 \\
Prop-FI & {80\%} & {-4023.7} & {95.35\%} & {16,000} & 80\% & 9e-4 \\
\hline
ELM & 20\% & -8111.4 & 47.3\% & 4000 & 20\% & 1e-4 \\
ELM & 40\% & -6795.55 & 56.45\% & 8000 & 40\% & 1e-4 \\
ELM & {60\%} & {-5601.55} & {68.49\%} & {12,000} & {60\%} & 1e-4 \\
ELM & {80\%} & {-4431.65} & {86.57\%} & {16,000} & 80\% & 1e-4 \\
\hline
\textbf{ApproxML} & \textbf{60\%} & \textbf{-3904.65} & \textbf{98.25\%} & \textbf{12,022} & \textbf{60.11\%} & \textbf{0.13} \\
\textbf{ApproxML} & \textbf{80\%} & \textbf{-3803.55} & \textbf{100.86\%} & \textbf{16,006} & \textbf{80.03\%} & \textbf{0.46} \\
\hline
GA & & -3836.4 & 100\% & 20,000 & 100\% \\
\end{tabular}}}
     \hfill
    \resizebox{0.48\textwidth}{!}{%
    \subfloat[\centering Frozen Lake]{\begin{tabular}{ccccccc}
          & Sample & Absolute & Relative & Absolute & Relative &  \\
  Method  & rate & fitness & fitness & computations & computations & p-value \\
\hline
HEA/FA & 20\% & 458.9 & 36.6\% & 1456.95 & 29.14\% & 1e-4  \\
HEA/FA & {40\%} & {600.35} & {47.88\%} & {2432.6} & {48.65\%} & {1e-4}  \\
HEA/FA & {60\%} & {680.4} & {54.27\%} & {3445.35} & {68.91\%} & {1e-4} \\
HEA/FA & {80\%} & {813} & {64.85\%} & {4404.9} & {88.1\%} & {1e-4} \\
\hline
Avg-FI & 20\% & 1005.35 & 80.19\% & 1000 & 20\% & 1e-4 \\
Avg-FI & 40\% & 1103.8 & 88.04\% & 2000 & 40\% & 1e-4 \\
Avg-FI & {60\%} & {1187.65} & {94.73\%} & {3000} & {60\%} & 4.7e-3 \\
\textbf{Avg-FI} & \textbf{80\%} & \textbf{1238.55} & \textbf{98.79\%} & \textbf{4000} & \textbf{80\%} & \textbf{0.39} \\
\hline
Prop-FI & 20\% & 938.3 & 74.84\% & 1000 & 20\% & 1e-4 \\
Prop-FI & 40\% & 1077.2 & 85.92\% & 2000 & 40\% & 1e-4 \\
Prop-FI & {60\%} & {1171.7} & {93.46\%} & {3000} & {60\%} & 1e-3 \\
\textbf{Prop-FI} & \textbf{80\%} & \textbf{1218.4} & \textbf{97.18\%} & \textbf{4000} & \textbf{80\%} & \textbf{0.07} \\
\hline
ELM & 20\% & 856.8 & 68.34\% & 1000 & 20\% & 1e-4  \\
ELM & {40\%} & {1052.5} & {83.95\%} & {2000} & {40\%} & {1e-4}  \\
ELM & {60\%} & {1174.4} & {93.67\%} & {3000} & {60\%} & {1e-4} \\
{ELM} & {80\%} & {1191.9} & {95.07\%} & {4000} & {80\%} & {0.007} \\
\hline
\textbf{ApproxML} & \textbf{60\%} & \textbf{1248.9} & \textbf{99.61\%} & \textbf{4730} & \textbf{94.6\%} & \textbf{0.8} \\
\textbf{ApproxML} & \textbf{80\%} & \textbf{1262.95} & \textbf{100.73\%} & \textbf{4892} & \textbf{97.84\%} & \textbf{0.58} \\
\hline
GA & & 1253.75 & 100\% & 5000 & 100\%  \\
\end{tabular}}}

    \resizebox{0.48\textwidth}{!}{%
    \subfloat[\centering Monster Cliff Walking]{\begin{tabular}{ccccccc}
          & Sample & Absolute & Relative & Absolute & Relative &  \\
  Method  & rate & fitness & fitness & computations & computations & p-value \\
\hline
HEA/FA & 20\% & -283.43 & 37.94\% & 2523.6 & 25.24\% & 1e-4  \\
HEA/FA & {40\%} & {-258.61} & {41.58\%} & {4872.8} & {48.73\%} & {1e-4}  \\
HEA/FA & {60\%} & {-223.3} & {48.15\%} & {6977.15} & {69.77\%} & {1e-4} \\
HEA/FA & {80\%} & {-219.47} & {48.99\%} & {9097.85} & {90.98\%} & {1e-4} \\
\hline
Avg-FI & 20\% & -217.27 & 49.49\% & 2000 & 20\% & 1e-4 \\
Avg-FI & 40\% & -163.40 & 65.81\% & 4000 & 40\% & 1e-4 \\
{Avg-FI} & {60\%} & {-131.62} & {81.69\%} & {6000} & {60\%} & {1e-4} \\
\textbf{Avg-FI} & \textbf{80\%} & \textbf{-107.34} & \textbf{100.17\%} & \textbf{8000} & \textbf{80\%} & \textbf{0.94} \\
\hline
Prop-FI & 20\% & -210.50 & 51.08\% & 2000 & 20\% & 1e-4 \\
Prop-FI & 40\% & -163.79 & 65.65\% & 4000 & 40\% & 1e-4 \\
Prop-FI & {60\%} & {-129.08} & {83.3\%} & {6000} & {60\%} & 1e-4 \\
\textbf{Prop-FI} & \textbf{80\%} & \textbf{-105.01} & \textbf{102.4\%} & \textbf{8000} & \textbf{80\%} & \textbf{0.33} \\
\hline
ELM & 20\% & -329.23 & 32.66\% & 2000 & 20\% & 1e-4  \\
ELM & {40\%} & {-243.25} & {44.20\%} & {4000} & {40\%} & {1e-4}  \\
ELM & {60\%} & {-166.76} & {64.48\%} & {6000} & {60\%} & {1e-4} \\
{ELM} & {80\%} & {-116.43} & {92.36\%} & {8000} & {80\%} & {0.005} \\
\hline
{ApproxML} & {60\%} & {-115.55} & {93.06\%} & {6336} & {63.36\%} & {0.02} \\
\textbf{ApproxML} & \textbf{80\%} & \textbf{-110.41} & \textbf{97.39\%} & \textbf{8170} & \textbf{81.7\%} & \textbf{0.24} \\
\hline
GA & & -107.53 & 100\% & 10,000 & 100\%  \\
\end{tabular}}}

\end{table*}

\section{Extensions}
\label{sec:extensions}

\subsection{Novelty search}
As mentioned in \autoref{sec:advantages-limitations}, the model should cover a large volume of the search space to generalize well to new individuals created by the genetic operators throughout the evolutionary run. We tested an alternative method to initialize the initial population. 
Instead of randomly generating the initial population, we perform Novelty Search \citep{lehman2011abandoning}.

In novelty search, fitness is abandoned in favor of finding distinct genomes, the idea being to keep vectors that are most distant than their $n$ nearest neighbors in the population (we used $n=20$).
This causes the individuals to be relatively far from each other at the end of novelty search, and thus cover a larger volume of the search space.

After this new initialization procedure the algorithm behaves exactly as in \autoref{sec:method}.

The results, shown in \autoref{tab:novelty}, indicate a minimal difference in fitness scores (cf. \autoref{tab:perm-tests}) when using this approach, except for statistical insignificance at the 60\% sample rate for Monster Cliff Walking, unlike with random initialization (\autoref{tab:perm-tests}). However, this initialization may be useful in other domains, possibly with larger search spaces. Alternatively, other exploration methods can be attempted within this context, e.g., Quality-Diversity \citep{pugh2016quality}.

\begin{table}
    \centering
    \caption{Initialization through novelty search: Results.}
    \label{tab:novelty}
\resizebox{0.47\textwidth}{!}{
    \subfloat[\centering Blackjack]{\begin{tabular}{cccccc}
Sample & Absolute & Relative & Absolute & Relative & p-value \\ rate & fitness & fitness & computations & computations\\
\hline
20\% & -4495.55 & 85.38\% & 4192 & 20.96\% & 1e-4 \\
40\% & -4009.95 & 95.68\% & 8066 & 40.33\% & 9e-4 \\
\textbf{60\%} & \textbf{-3877.7} & \textbf{98.93\%} & \textbf{12,026} & \textbf{60.13\%} & \textbf{0.35} \\
\textbf{80\%} & \textbf{-3848.75} & \textbf{99.67\%} & \textbf{16,013} & \textbf{80.07\%} & \textbf{0.81} \\
GA & -3836.4 & 100\% & 20,000 & 100\% \\
\end{tabular}}}
    
    \resizebox{0.47\textwidth}{!}{
    \subfloat[\centering Frozen Lake]{\begin{tabular}{cccccc}
Sample & Absolute & Relative & Absolute     & Relative     & p-value \\ 
rate   & fitness  & fitness  & computations & computations  \\
\hline
20\%          & 1159             & 92.44\%            & 2836           & 56.72\%          & 3e-4  \\
\textbf{40\%} & \textbf{1254.65} & \textbf{100\%}    & \textbf{4295}  & \textbf{85.9\%}  & \textbf{0.97}  \\
\textbf{60\%} & \textbf{1241.95} & \textbf{99.06\%}  & \textbf{4700}  & \textbf{94\%}    & \textbf{0.6} \\
\textbf{80\%} & \textbf{1231.4}  & \textbf{98.22\%}  & \textbf{4882}  & \textbf{97.64\%} & \textbf{0.26} \\
GA            & 1253.75          & 100\%             & 5000           & 100\%            & \\
\end{tabular}}}

\resizebox{0.47\textwidth}{!}{
    \subfloat[\centering Monster Cliff Walking]{\begin{tabular}{cccccc}
Sample & Absolute & Relative & Absolute & Relative & p-value \\ rate & fitness & fitness & computations & computations\\
\hline
{20\%} & {-129.04} & {83.33\%} & {4180} & {41.8\%} & {1e-4} \\
{40\%} & {-130.31} & {82.52\%} & {4738} & {47.38\%} & {1e-4} \\
\textbf{60\%} & \textbf{-112.67} & \textbf{95.43\%} & \textbf{6340} & \textbf{63.4\%} & \textbf{0.12} \\
\textbf{80\%} & \textbf{-109.59} & \textbf{98.12\%} & \textbf{8165} & \textbf{81.65\%} & \textbf{0.47} \\
GA & -14.71 & 100\% & 10,000 & 100\% \\
\end{tabular}}}
\end{table}

\subsection{Hidden actual fitness scores in prediction mode}
In existing fitness-approximation methods, as discussed in \autoref{sec:prev}, some individuals in the population have their fitness values approximated, while the rest receive actual fitness scores. Our preliminary investigation, which employed this scheme, suggested that individuals with actual fitness scores might ``hijack'' the evolutionary process. In cases of pessimistic approximations, the individuals with actual fitness scores are likely to be selected, to then repeatedly reproduce, thus restricting the exploration of the search space \citep{jin2005comprehensive}.

In evolution mode of our baseline method, the actual fitness scores are both given to the GA and added to the ML dataset.
Another approach is to hide actual fitness scores from the GA in prediction mode, only adding them to the ML dataset. The sampled individuals still receive approximate fitness scores, even though their actual fitness scores are computed.

This latter approach harms the accuracy of the fitness scores since the approximate fitness scores are usually less accurate than the actual fitness scores. On the other hand, this approach prevents the potential problem described above. In contrast to the baseline method, the calculation of the actual fitness scores and model training in prediction mode is independent of the evolutionary process when true fitness scores are hidden from the GA. As a result, fitness computation and model training can be executed in a separate process in parallel to the evolution, significantly reducing runtime. We plan to perform concrete experiments on this approach in the future.

\section{Concluding Remarks and Future Work}
\label{sec:conclude}
This paper presented a generic method to integrate machine learning models in fitness approximation. Our approach is useful for domains wherein the fitness score calculation is computationally expensive, such as running a simulator. We used Gymnasium simulators for evaluating actual fitness scores, and Ridge and Lasso models for learning the fitness-approximation models. 

Our analysis includes a rigorous comparison between different methods for: 
1) switching between actual and approximate fitness, 
2) sampling the population, and
3) weighting the samples. 

\textit{Our results show a significant reduction in GA runtime, with a small price in fitness for low sample rates, and no price for high sample rates.}

Further enhancements can be incorporated into our method by employing more complex ML models, such as Random Forest \citep{breiman2001random}, XGBoost \citep{chen2016xgboost}, or Deep Networks \citep{prince2023understanding}. While these models have the potential to improve fitness approximation, it is worth noting that they are typically computationally intensive and may not be suitable for domains with limited fitness computation time.

Additionally, our method can be refined by leveraging domain-specific knowledge or additional data science concepts (under-sampling \citep{liu2008exploratory}, over-sampling \citep{chawla2002smote}, feature engineering \citep{zheng2018feature}, etc.), to improve the generality of the population dataset and, consequently, the accuracy of the model. These approaches have the potential to enhance the overall performance of our solution.

\section*{Acknowledgments}
This research was partially supported by the following grants:
grant \#2714/19 from the Israeli Science Foundation; 
Israeli Smart Transportation Research Center (ISTRC); 
Israeli Council for Higher Education (CHE) via the Data Science Research Center, Ben-Gurion University of the Negev, Israel.

\small
\bibliography{references.bib}
\bibliographystyle{abbrvnat}

\end{document}